\pdfoutput=1
\documentclass[11pt]{article}

\usepackage[preprint]{acl}

\usepackage{times}
\usepackage{latexsym}
\usepackage{todonotes}
\usepackage{utfsym}
\usepackage[T1]{fontenc}
\usepackage[utf8]{inputenc}

\usepackage{microtype}

\usepackage{inconsolata}

\usepackage{graphicx}

\usepackage{amssymb}

\title{Audio-visual training for improved grounding in video-text LLMs}

\author{Shivprasad Sagare \and Hemachandran S \and Kinshuk Sarabhai \and Prashant Ullegaddi \and Rajeshkumar SA \\
Phronetic AI}
\author{
 \textbf{Shivprasad Sagare},
 \textbf{Hemachandran S.},
 \textbf{Kinshuk Sarabhai},
 \\
 \textbf{Prashant Ullegaddi},
 \textbf{Rajeshkumar SA}
 \\
    PhroneticAI
\\
 \small{
   \textbf{Correspondence:} \href{mailto:shivprasad.sagare@phronetic.ai}{shivprasad.sagare@phronetic.ai}
 }
}
\begin{document}
\maketitle
\begin{abstract}
Recent advances in multimodal LLMs, have led to several video-text models being proposed for critical video-related tasks. However, most of the previous works support visual input only, essentially muting the audio signal in the video. Few models that support both audio and visual input, are not explicitly trained on audio data. Hence, the effect of audio towards video understanding is largely unexplored. To this end, we propose a model architecture that handles audio-visual inputs explicitly. We train our model with both audio and visual data from a video instruction-tuning dataset. Comparison with vision-only baselines, and other audio-visual models showcase that training on audio data indeed leads to improved grounding of responses. For better evaluation of audio-visual models, we also release a human-annotated benchmark dataset, with audio-aware question-answer pairs.
\end{abstract}

\section{Introduction}
Conversational agents fueled by LLMs have made it possible for us to interact in a new way with data from multiple modalities\cite{yin2024survey}\cite{wadekar2024evolution}. Image-text multimodal LLMs(MLLMs) like LLaVA\cite{liu2023visual} have demonstrated the effectiveness of visual instruction-tuning(IT) data. Several works like VideoChatGPT\cite{maaz2023videochatgpt}, VideoChat\cite{li2024videochat}, PLLaVa\cite{xu2024pllava} have extended the image-text model architecture for video related tasks.

\begin{figure}[t]
  \includegraphics[width=\columnwidth]{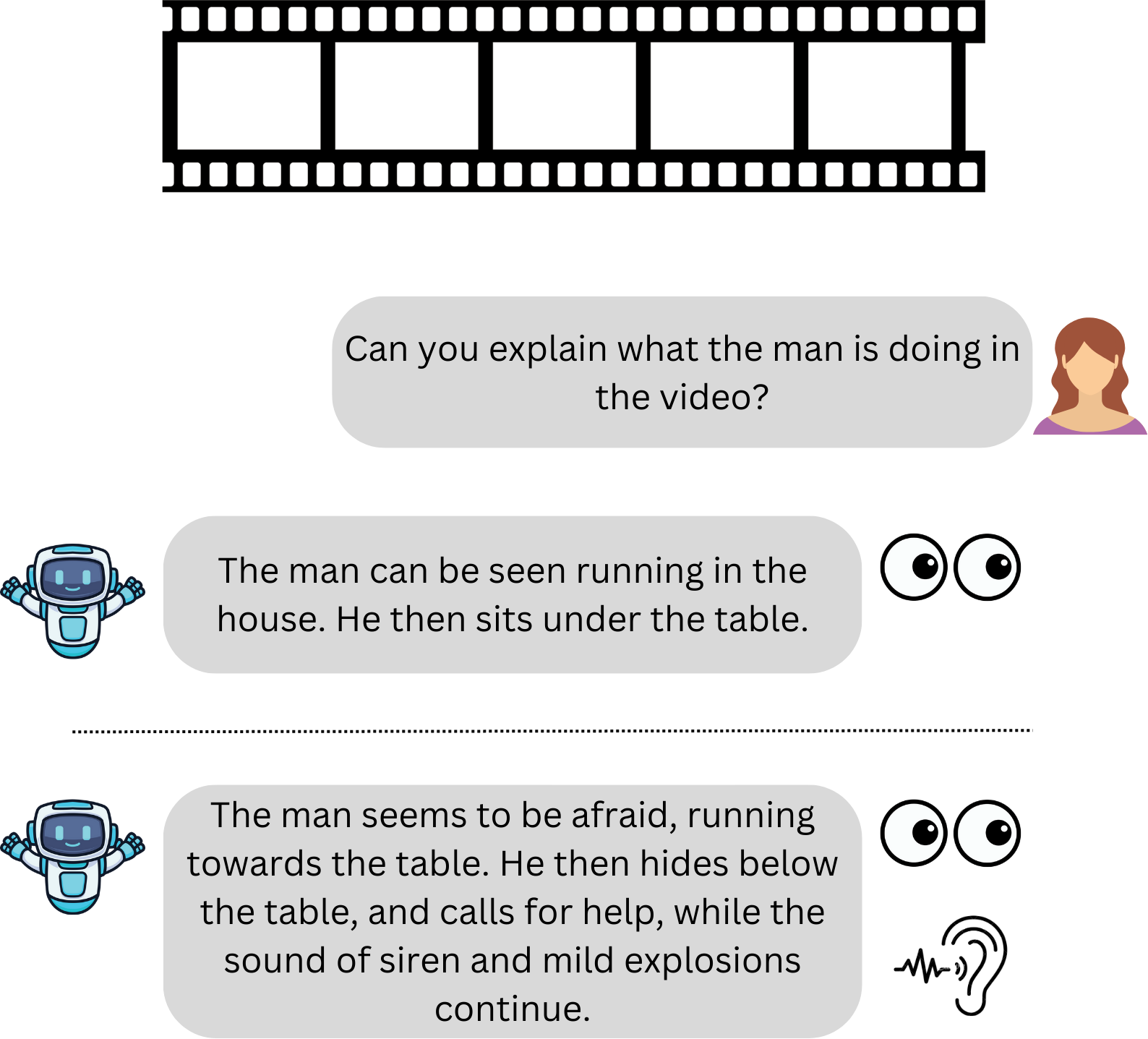}
  \caption{An example of improved grounding in the video-text LLM outputs, due to the additional audio signal as input.}
  \label{fig:task_example}
\end{figure}

However, most of the above works rely only on the visual input, and do not consider audio signal for video understanding. In real world, listening to audio while playing the video, adds immensely to our perception of the video. We propose a video-text MLLM, with Phi-2\cite{gunasekar2023textbooks} as the LLM backbone. It supports both audio and visual inputs, using Whisper\cite{radford2022robust} and sigLIP\cite{zhai2023sigmoid} encoders respectively. Unlike previous works, we train the model using audio data explicitly, in addition to the visual data. We aim to explore the role of audio in video understanding and if audio input can be utilized for better grounding of video-text LLMs. We also explore the creation of better benchmarks that encompass variety of question-answer pairs. Evaluation on several benchmarks demonstrates the effectiveness of audio as an additional signal in better understanding of the video content.

Overall we make the following key contributions: 
\textbf{1}.We propose an efficient video-text MLLM architecture consisting of separate encoders to process the audio and visual inputs. 
\newline
\textbf{2}.We train our video-text model using both audio and visual signals simultaneously, aiming to explore the effect of audio input on model outputs.
\newline
\textbf{3}.We release a human-annotated benchmark dataset containing video instruction-tuning samples, which are audio-aware.

% \begin{table}[]
%     \centering
%     \begin{tabular}{c|c|c|c}
%          Models & Visual & Audio & Audio-visual  \\
%          \hline
%          VideoChatGPT & \usym{1F5F8} & -- & --\\
%          LLaSM & -- & \usym{1F5F8} & --\\
%          Video-LLaMA & \usym{1F5F8} & \usym{2718} & \usym{2718}\\
%          NExT-GPT & \usym{1F5F8} & \usym{1F5F8} & \usym{2718}\\
%          our & \usym{1F5F8} & \usym{1F5F8} & \usym{1F5F8}
%     \end{tabular}
%     \caption{Comparing MLLMs based on the input modalities supported, and the training data. -- indicates that the input modality isn't supported. \usym{2718} indicates that the input modality is supported, but the model isn't trained using such data. \usym{1F5F8} indicates that the model architecture supports the input modality, and has also been explicitly trained on such data.}
%     \label{tab:related_work}
% \end{table}

\begin{table}[]
    \centering
    \begin{tabular}{c|c|c|c}
         Models & Visual & Audio & Audio-visual  \\
         \hline
         VideoChatGPT & \checkmark & -- & --\\
         LLaSM & -- & \checkmark & --\\
         Video-LLaMA & \checkmark & $\times$ & $\times$\\
         NExT-GPT & \checkmark & \checkmark & $\times$\\
         our & \checkmark & \checkmark & \checkmark
    \end{tabular}
    \caption{Comparing MLLMs based on the input modalities supported, and the training data. -- indicates that the input modality isn't supported. $\times$ indicates that the input modality is supported, but the model isn't trained using such data. \checkmark indicates that the model architecture supports the input modality, and has also been explicitly trained on such data.}
    \label{tab:related_work}
\end{table}

\section{Related work}
\label{sec:relatedWork}
\noindent\underline{\textbf{Vision-text MLLMs}}: 
LLaVA\cite{liu2023visual}, MiniGPT4\cite{zhu2023minigpt4} have showcased the efficacy of visual instruction-tuning datasets for image-text tasks. Bunny\cite{he2024efficient} explores a similar idea but using lightweight LLM backbones. Several works like PLLaVA\cite{xu2024pllava} build on the top of image-text MLLMs to support video input. VideoChatGPT\cite{maaz2023videochatgpt} extends the CLIP image encoder\cite{radford2021learning} to videos by averaging the representations across spatial and temporal dimensions.

\noindent\underline{\textbf{Audio-text MLLMs}}:
Similar to vision-text, there has been recent work in fusing audio input features with text LLM for several audio-text tasks\cite{zhang2023speechgpt}. LLaSM\cite{shu2023llasm} demonstrates the effectiveness of pretraining the projector layers using speech-to-text data. Some previous works like AudioGPT\cite{huang2023audiogpt} build on LLM-based planning and tool-use to solve several audio tasks at once.

\noindent\underline{\textbf{Audio-vision-text MLLMs}}
Similar to our work, Video-LLaMA\cite{zhang2023videollama}, and NExT-GPT\cite{wu2023nextgpt} support audio and visual input simultaneously, both relying on unified modality encoder ImageBind\cite{girdhar2023imagebind}. However, Video-LLaMA is trained only on visual IT datasets, assuming the audio branch learns implicitly. NExT-GPT is trained using cross-modal IT dataset, but doesn't utilize audio-visual simultaneous input from videos. Unlike previous works, we explore training using audio-visual input from videos simultaneously, and explore the grounding effect it has on model outputs.

\section{Model architecture}
Following the idea of fusing the modality inputs into LLM\cite{liu2023visual}\cite{zhang2023videollama}, we build a video-text MLLM architecture consisting of two separate branches for audio and visual inputs. Each branch consists of modality encoder, projector layers to transform the encoder representations into LLM embedding space, followed by the backbone LLM. 

We use Whisper\cite{radford2022robust} as an audio encoder, and use its last hidden state as audio representations\cite{shu2023llasm}. To encode the video, we use sigLIP image encoder\cite{zhai2023sigmoid}. Following \cite{maaz2023videochatgpt}, we treat video as a sequence of images, and compute frame representations using sigLIP. We then compute spatial and temporal average of representations across 100 uniformly sampled frames, and use it as a video representation. Inspired from Bunny\cite{he2024efficient}, we rely on low-cost, efficient, lightweight LLM backbone with 2.7 Billion parameters, phi-2\cite{gunasekar2023textbooks}. Projector layer for both vision and audio branch is mlp2x-gelu\cite{he2024efficient}. 

\begin{figure*}[]
  \includegraphics[width=\textwidth]{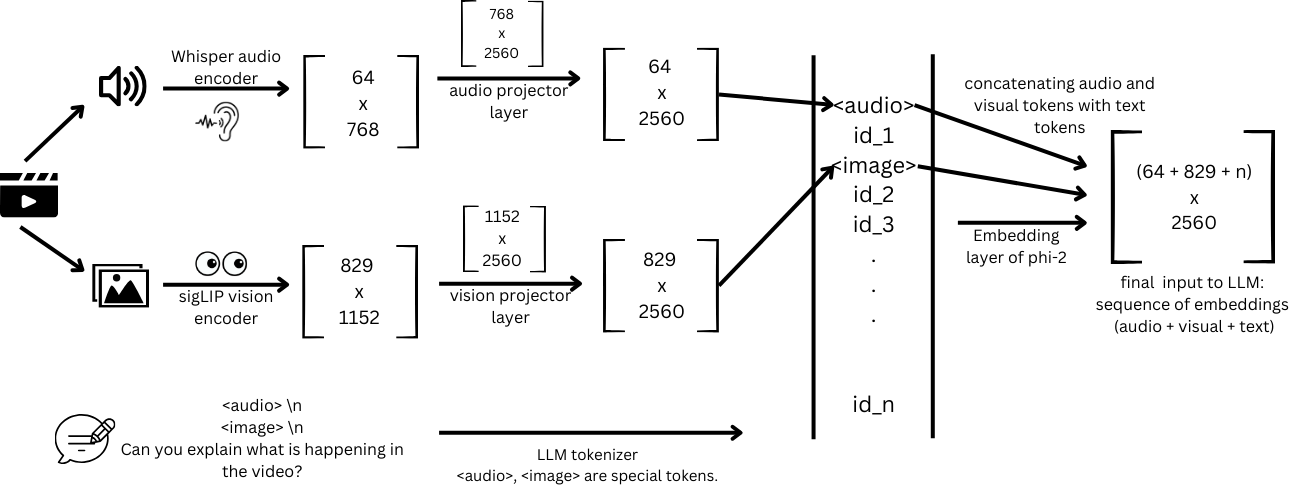}
  \caption{Tensor dimensions in the figure denote the flow of data through the encoder and projector layers. Audio encoder(Whisper) and video encoder(using sigLIP) produce 64 and 829 token embeddings respectively, which are then concatenated with the text token embeddings as the final input to the LLM. Unlike previous works, we train both the audio and vision branch simultaneously using a video instruction tuning dataset.}
  \label{fig:model_architecture}
\end{figure*}

The exact flow of input data through both the audio and visual branches is shown in the form of tensor dimensions, in figure \ref{fig:model_architecture}. Audio and visual input is converted into 64 and 829 token embeddings respectively. Audio, visual, and text token embeddings are then concatenated before passing to the backbone LLM.

% For instruction tuning the model, we follow the template as shown in the figure \ref{fig:experiments}. <audio> and <image> are placeholder tokens, that are later substituted by the respective input modality token embeddings.

\section{Training setup and datasets}
Training different components of our model with appropriate data is a key focus of our research. Typically, these MLLMs go through a pretraining stage, followed by the finetuning stage.

\noindent\underline{\textbf{Pretraining}}:
Pretraining aims to align different modalities to text LLM space, by training on some generic modality-to-text task. Only projector layer weights are trained during this phase, while encoders, and LLM weights are frozen.
We pretrain our audio projector layers using a combination of Speech-to-Text(STT) dataset(CommonVoice\cite{ardila-etal-2020-common}) and audio captioning dataset(AudioCaps\cite{kim-etal-2019-audiocaps}) with 50K samples each. We convert these datasets into our instruction-tuning prompt template by creating 10 instructions each for transcription and captioning.
Since our visual branch relies on image encoder, we employ already trained checkpoint by Bunny\cite{he2024efficient} to initialize vision projector layers. It has been trained on 2M subset of an image-text dataset LAION\cite{schuhmann2022laionb}. We freeze the vision branch while pretraining audio projector layers, and vice versa.

\noindent\underline{\textbf{Finetuning}}:
Finetuning or instruction tuning is aimed to train the LLM model to follow the exact requests or questions in the user prompt\cite{ouyang2022training}. Unlike previous works, we explicitly train both the audio and visual branches of the model simultaneously, using video instruction-tuning dataset containing both the audio and visual data. We rely on VideoInstruct100K\cite{maaz2023videochatgpt} dataset with 100K samples containing video and question answer pair. Although the dataset authors had used the dataset only for visual instruction tuning, we extract the audios(wav format) from the videos(mp4 format) for our use-case. 

We aim to explore if including audio features during training helps the model to better understand the video. To measure this effect, we also train a baseline vision-only model, without the audio branch. We train the vision branch of the model, using the visual data from same dataset.

\noindent\underline{\textbf{Experiment details}}
We implement the audio and video functionality by extending the codebases of Bunny and LLaSM. We use Whisper-small, siglip-so400m-patch14-384, and phi-2 models from HuggingFace. Pretraining for audio projector layer was done using A100, with global batch size of 128. Finetuning was implemented using LoRA for training LLM weights, on A40 machine.
\begin{table*}[]
\centering
\begin{tabular}{|c|c|c|c|}
  \hline
  Metrics &  visual-only model (our) & video-llama & audio-visual model (our) \\
  \hline
  Correctness of Information & 2.34 & 1.96 & \textbf{2.69} \\
  \hline
  Detail Orientation & 2.35 & 2.18 & \textbf{2.49}\\
  \hline
  Contextual Understanding & 2.74 & 2.16 & \textbf{3.04}\\
  \hline
  Temporal Understanding & 1.97 & 1.82 & \textbf{2.22}\\
  \hline
  Consistency & 2.45 & 1.79 & \textbf{2.71} \\
  \hline
  Average & 2.37 & 1.98 & \textbf{2.63} \\
  \hline
\end{tabular}
\caption{Results on VideoChatGPT evaluation framework. Our audio-visual training setup shows impressive results when compared with other audio-vision model(Video-LLaMA), as well our vision-only baseline.}
\label{tab:videochatgpt_dataset_evaluation}
\end{table*}

\begin{table*}[]
\centering
\begin{tabular}{|c|c|c|c|}
  \hline
  Metrics &  visual-only model (our) & video-llama & audio-visual model (our) \\
  \hline
  Correctness of Information & 2.34 & 1.49 & \textbf{2.77} \\
  \hline
  Detail Orientation & 2.36 & 1.7 & \textbf{2.44} \\
  \hline
  Contextual Understanding & 2.75 & 1.92 & \textbf{3.04} \\
  \hline
  Temporal Understanding & 2.17 & 1.4 & \textbf{2.4} \\
  \hline
  Average & 2.40 & 1.62 & \textbf{2.66} \\
  \hline
\end{tabular}
\caption{Results on our benchmark dataset. Results illustrate similar trend as above, where training on audio signals helps the model to generate more accurate responses. We haven't yet incorporated evaluation for consistency metric in our benchmark dataset.}
\label{tab:our_benchmark_dataset_evaluation}
\end{table*}

\section{Benchmark dataset}
Several evaluation criteria and datasets have been introduced to benchmark the vision-text MLLMs\cite{chen-dolan-2011-collecting}\cite{maaz2023videochatgpt}\cite{7298698}. VideoChatGPT has released a human verified benchmark dataset consisting of 500 videos and corresponding question-answer pairs for video-text tasks. However, these benchmarks do not consider audio information while creating the question-answer pairs based on videos. Thus, it is challenging to evaluate the capability of model to attend to both the audio and visual signals while generating the output.

Therefore, we annotate such an audio-visual instruction-tuning dataset that contains question-answer pairs based both on audio and visual information in the video. We include both generic questions, like 'What is happening in the video?', as well as more specific questions related to the video. Answer of each question is around 2 sentences, with most of the videos available on YouTube. We release a set of 120 such samples, as we intend to scale the size and quality of the data in future. Example samples from our benchmark dataset are shown below.\\

\noindent\underline{\textbf{Sample 1}} \\
\textbf{Question}: What is the man doing in the video? \\
\textbf{Answer}: In the video, the man fires his gun upwards, producing the sharp sound of a bullet being shot. The echo reverberates through the air, adding tension and intensity to the scene.
\\
\newpage
\noindent\underline{\textbf{Sample 2}} \\
\textbf{Question}: What is the man on the stage mentoring about in the video? \\
\textbf{Answer}: The workshop leader, mentors a student on speaking louder for clarity. He asks the student to raise the volume from level 3 to level 7. Finally, the student earns an applause from the audience in the communication workshop.

\section{Evaluation}
We extensively evaluate our model using VideoChatGPT evaluation framework across 5 key metrics. It relies on LLM-based evaluation(using GPT-3.5) which rates the output on the scale of 1-5. We compare our audio-visual model with the visual-only baseline that we have trained, as well as other audio-visual model, Video-LLaMA. The evaluation results are summarized in the table \ref{tab:videochatgpt_dataset_evaluation}. Similarly, we evaluate on our benchmark dataset, and observe the similar trends, as summarized in \ref{tab:our_benchmark_dataset_evaluation}. 

The audio-visual model clearly performs better than the vision-only baseline by a margin. Interestingly, Video-LLaMA which is also an audio-visual model performs poorly on both the benchmarks. Video-LLaMA does not utilize the audio inputs explicitly, and instead rely on visual signals only during training. We could not compare against another audio-visual model, NExT-GPT, as it relies on LLaMA-v0 weights which couldn't be available to us due to licensing. 

Qualitative analysis of audio-visual model outputs demonstrate better overall quality compared to vision-only model. We also analyze the model outputs at intermediate stages, i.e. after pre-training. Our model could very well generate the captions of audio data, which showed the efficacy of pretraining step. There is scope for better encoding strategies and training regimes for utilizing audio information even more. 

\section{Conclusion and future work}
We performed several experiments and evaluations to specifically study how audio signal can be utilized for better video understanding. Training the MLLM simultaneously on audio-visual signals of the video indeed results in a better performance, as seen in quantitative evaluation using several metrics. We also contributed a benchmark dataset curated to evaluate the video-understanding capability using both visual and audio information. 

Based on these results, we are motivated to experiment with sophisticated ways of incorporating audio and visual signals together for video related tasks. Future work also consists of the extensive analysis of the type of question-answer pairs in video IT datasets, and work on creating better evaluation benchmarks catering to wide range of video-related use-cases.   

\bibliography{custom}

\begin{thebibliography}{24}
\providecommand{\natexlab}[1]{#1}

\bibitem[{Ardila et~al.(2020)Ardila, Branson, Davis, Kohler, Meyer, Henretty, Morais, Saunders, Tyers, and Weber}]{ardila-etal-2020-common}
Rosana Ardila, Megan Branson, Kelly Davis, Michael Kohler, Josh Meyer, Michael Henretty, Reuben Morais, Lindsay Saunders, Francis Tyers, and Gregor Weber. 2020.
\newblock \href {https://aclanthology.org/2020.lrec-1.520} {Common voice: A massively-multilingual speech corpus}.
\newblock In \emph{Proceedings of the Twelfth Language Resources and Evaluation Conference}, pages 4218--4222, Marseille, France. European Language Resources Association.

\bibitem[{Chen and Dolan(2011)}]{chen-dolan-2011-collecting}
David Chen and William Dolan. 2011.
\newblock \href {https://aclanthology.org/P11-1020} {Collecting highly parallel data for paraphrase evaluation}.
\newblock In \emph{Proceedings of the 49th Annual Meeting of the Association for Computational Linguistics: Human Language Technologies}, pages 190--200, Portland, Oregon, USA. Association for Computational Linguistics.

\bibitem[{Girdhar et~al.(2023)Girdhar, El-Nouby, Liu, Singh, Alwala, Joulin, and Misra}]{girdhar2023imagebind}
Rohit Girdhar, Alaaeldin El-Nouby, Zhuang Liu, Mannat Singh, Kalyan~Vasudev Alwala, Armand Joulin, and Ishan Misra. 2023.
\newblock \href {https://arxiv.org/abs/2305.05665} {Imagebind: One embedding space to bind them all}.
\newblock \emph{Preprint}, arXiv:2305.05665.

\bibitem[{Gunasekar et~al.(2023)Gunasekar, Zhang, Aneja, Cesar, Mendes, Giorno, Gopi, Javaheripi, Kauffmann, de~Rosa, Saarikivi, Salim, Shah, Singh~Behl, Wang, Bubeck, Eldan, Kalai, Lee, and Li}]{gunasekar2023textbooks}
Suriya Gunasekar, Yi~Zhang, Jyoti Aneja, Caio Cesar, Teodoro Mendes, Allie~Del Giorno, Sivakanth Gopi, Mojan Javaheripi, Piero Kauffmann, Gustavo de~Rosa, Olli Saarikivi, Adil Salim, Shital Shah, Harkirat Singh~Behl, Xin Wang, Sébastien Bubeck, Ronen Eldan, Adam~Tauman Kalai, Yin~Tat Lee, and Yuanzhi Li. 2023.
\newblock \href {https://www.microsoft.com/en-us/research/publication/textbooks-are-all-you-need/} {Textbooks are all you need}.

\bibitem[{He et~al.(2024)He, Liu, Wu, Yuan, Wang, Huang, and Zhao}]{he2024efficient}
Muyang He, Yexin Liu, Boya Wu, Jianhao Yuan, Yueze Wang, Tiejun Huang, and Bo~Zhao. 2024.
\newblock \href {https://arxiv.org/abs/2402.11530} {Efficient multimodal learning from data-centric perspective}.
\newblock \emph{Preprint}, arXiv:2402.11530.

\bibitem[{Heilbron et~al.(2015)Heilbron, Escorcia, Ghanem, and Niebles}]{7298698}
Fabian~Caba Heilbron, Victor Escorcia, Bernard Ghanem, and Juan~Carlos Niebles. 2015.
\newblock \href {https://doi.org/10.1109/CVPR.2015.7298698} {Activitynet: A large-scale video benchmark for human activity understanding}.
\newblock In \emph{2015 IEEE Conference on Computer Vision and Pattern Recognition (CVPR)}, pages 961--970.

\bibitem[{Huang et~al.(2023)Huang, Li, Yang, Shi, Chang, Ye, Wu, Hong, Huang, Liu, Ren, Zhao, and Watanabe}]{huang2023audiogpt}
Rongjie Huang, Mingze Li, Dongchao Yang, Jiatong Shi, Xuankai Chang, Zhenhui Ye, Yuning Wu, Zhiqing Hong, Jiawei Huang, Jinglin Liu, Yi~Ren, Zhou Zhao, and Shinji Watanabe. 2023.
\newblock \href {https://arxiv.org/abs/2304.12995} {Audiogpt: Understanding and generating speech, music, sound, and talking head}.
\newblock \emph{Preprint}, arXiv:2304.12995.

\bibitem[{Kim et~al.(2019)Kim, Kim, Lee, and Kim}]{kim-etal-2019-audiocaps}
Chris~Dongjoo Kim, Byeongchang Kim, Hyunmin Lee, and Gunhee Kim. 2019.
\newblock \href {https://doi.org/10.18653/v1/N19-1011} {{A}udio{C}aps: Generating captions for audios in the wild}.
\newblock In \emph{Proceedings of the 2019 Conference of the North {A}merican Chapter of the Association for Computational Linguistics: Human Language Technologies, Volume 1 (Long and Short Papers)}, pages 119--132, Minneapolis, Minnesota. Association for Computational Linguistics.

\bibitem[{Li et~al.(2024)Li, He, Wang, Li, Wang, Luo, Wang, Wang, and Qiao}]{li2024videochat}
KunChang Li, Yinan He, Yi~Wang, Yizhuo Li, Wenhai Wang, Ping Luo, Yali Wang, Limin Wang, and Yu~Qiao. 2024.
\newblock \href {https://arxiv.org/abs/2305.06355} {Videochat: Chat-centric video understanding}.
\newblock \emph{Preprint}, arXiv:2305.06355.

\bibitem[{Liu et~al.(2023)Liu, Li, Wu, and Lee}]{liu2023visual}
Haotian Liu, Chunyuan Li, Qingyang Wu, and Yong~Jae Lee. 2023.
\newblock \href {https://arxiv.org/abs/2304.08485} {Visual instruction tuning}.
\newblock \emph{Preprint}, arXiv:2304.08485.

\bibitem[{Maaz et~al.(2023)Maaz, Rasheed, Khan, and Khan}]{maaz2023videochatgpt}
Muhammad Maaz, Hanoona Rasheed, Salman Khan, and Fahad~Shahbaz Khan. 2023.
\newblock \href {https://arxiv.org/abs/2306.05424} {Video-chatgpt: Towards detailed video understanding via large vision and language models}.
\newblock \emph{Preprint}, arXiv:2306.05424.

\bibitem[{Ouyang et~al.(2022)Ouyang, Wu, Jiang, Almeida, Wainwright, Mishkin, Zhang, Agarwal, Slama, Ray, Schulman, Hilton, Kelton, Miller, Simens, Askell, Welinder, Christiano, Leike, and Lowe}]{ouyang2022training}
Long Ouyang, Jeff Wu, Xu~Jiang, Diogo Almeida, Carroll~L. Wainwright, Pamela Mishkin, Chong Zhang, Sandhini Agarwal, Katarina Slama, Alex Ray, John Schulman, Jacob Hilton, Fraser Kelton, Luke Miller, Maddie Simens, Amanda Askell, Peter Welinder, Paul Christiano, Jan Leike, and Ryan Lowe. 2022.
\newblock \href {https://arxiv.org/abs/2203.02155} {Training language models to follow instructions with human feedback}.
\newblock \emph{Preprint}, arXiv:2203.02155.

\bibitem[{Radford et~al.(2021)Radford, Kim, Hallacy, Ramesh, Goh, Agarwal, Sastry, Askell, Mishkin, Clark, Krueger, and Sutskever}]{radford2021learning}
Alec Radford, Jong~Wook Kim, Chris Hallacy, Aditya Ramesh, Gabriel Goh, Sandhini Agarwal, Girish Sastry, Amanda Askell, Pamela Mishkin, Jack Clark, Gretchen Krueger, and Ilya Sutskever. 2021.
\newblock \href {https://arxiv.org/abs/2103.00020} {Learning transferable visual models from natural language supervision}.
\newblock \emph{Preprint}, arXiv:2103.00020.

\bibitem[{Radford et~al.(2022)Radford, Kim, Xu, Brockman, McLeavey, and Sutskever}]{radford2022robust}
Alec Radford, Jong~Wook Kim, Tao Xu, Greg Brockman, Christine McLeavey, and Ilya Sutskever. 2022.
\newblock \href {https://arxiv.org/abs/2212.04356} {Robust speech recognition via large-scale weak supervision}.
\newblock \emph{Preprint}, arXiv:2212.04356.

\bibitem[{Schuhmann et~al.(2022)Schuhmann, Beaumont, Vencu, Gordon, Wightman, Cherti, Coombes, Katta, Mullis, Wortsman, Schramowski, Kundurthy, Crowson, Schmidt, Kaczmarczyk, and Jitsev}]{schuhmann2022laionb}
Christoph Schuhmann, Romain Beaumont, Richard Vencu, Cade~W Gordon, Ross Wightman, Mehdi Cherti, Theo Coombes, Aarush Katta, Clayton Mullis, Mitchell Wortsman, Patrick Schramowski, Srivatsa~R Kundurthy, Katherine Crowson, Ludwig Schmidt, Robert Kaczmarczyk, and Jenia Jitsev. 2022.
\newblock \href {https://openreview.net/forum?id=M3Y74vmsMcY} {{LAION}-5b: An open large-scale dataset for training next generation image-text models}.
\newblock In \emph{Thirty-sixth Conference on Neural Information Processing Systems Datasets and Benchmarks Track}.

\bibitem[{Shu et~al.(2023)Shu, Dong, Chen, Huang, Zhang, Shi, Xiang, and Shi}]{shu2023llasm}
Yu~Shu, Siwei Dong, Guangyao Chen, Wenhao Huang, Ruihua Zhang, Daochen Shi, Qiqi Xiang, and Yemin Shi. 2023.
\newblock \href {https://arxiv.org/abs/2308.15930} {Llasm: Large language and speech model}.
\newblock \emph{Preprint}, arXiv:2308.15930.

\bibitem[{Wadekar et~al.(2024)Wadekar, Chaurasia, Chadha, and Culurciello}]{wadekar2024evolution}
Shakti~N. Wadekar, Abhishek Chaurasia, Aman Chadha, and Eugenio Culurciello. 2024.
\newblock \href {https://arxiv.org/abs/2405.17927} {The evolution of multimodal model architectures}.
\newblock \emph{Preprint}, arXiv:2405.17927.

\bibitem[{Wu et~al.(2023)Wu, Fei, Qu, Ji, and Chua}]{wu2023nextgpt}
Shengqiong Wu, Hao Fei, Leigang Qu, Wei Ji, and Tat-Seng Chua. 2023.
\newblock \href {https://arxiv.org/abs/2309.05519} {Next-gpt: Any-to-any multimodal llm}.
\newblock \emph{Preprint}, arXiv:2309.05519.

\bibitem[{Xu et~al.(2024)Xu, Zhao, Zhou, Lin, Ng, and Feng}]{xu2024pllava}
Lin Xu, Yilin Zhao, Daquan Zhou, Zhijie Lin, See~Kiong Ng, and Jiashi Feng. 2024.
\newblock \href {https://arxiv.org/abs/2404.16994} {Pllava : Parameter-free llava extension from images to videos for video dense captioning}.
\newblock \emph{Preprint}, arXiv:2404.16994.

\bibitem[{Yin et~al.(2024)Yin, Fu, Zhao, Li, Sun, Xu, and Chen}]{yin2024survey}
Shukang Yin, Chaoyou Fu, Sirui Zhao, Ke~Li, Xing Sun, Tong Xu, and Enhong Chen. 2024.
\newblock \href {https://arxiv.org/abs/2306.13549} {A survey on multimodal large language models}.
\newblock \emph{Preprint}, arXiv:2306.13549.

\bibitem[{Zhai et~al.(2023)Zhai, Mustafa, Kolesnikov, and Beyer}]{zhai2023sigmoid}
Xiaohua Zhai, Basil Mustafa, Alexander Kolesnikov, and Lucas Beyer. 2023.
\newblock \href {https://arxiv.org/abs/2303.15343} {Sigmoid loss for language image pre-training}.
\newblock \emph{Preprint}, arXiv:2303.15343.

\bibitem[{Zhang et~al.(2023{\natexlab{a}})Zhang, Li, Zhang, Zhan, Wang, Zhou, and Qiu}]{zhang2023speechgpt}
Dong Zhang, Shimin Li, Xin Zhang, Jun Zhan, Pengyu Wang, Yaqian Zhou, and Xipeng Qiu. 2023{\natexlab{a}}.
\newblock \href {https://arxiv.org/abs/2305.11000} {Speechgpt: Empowering large language models with intrinsic cross-modal conversational abilities}.
\newblock \emph{Preprint}, arXiv:2305.11000.

\bibitem[{Zhang et~al.(2023{\natexlab{b}})Zhang, Li, and Bing}]{zhang2023videollama}
Hang Zhang, Xin Li, and Lidong Bing. 2023{\natexlab{b}}.
\newblock \href {https://arxiv.org/abs/2306.02858} {Video-llama: An instruction-tuned audio-visual language model for video understanding}.
\newblock \emph{Preprint}, arXiv:2306.02858.

\bibitem[{Zhu et~al.(2023)Zhu, Chen, Shen, Li, and Elhoseiny}]{zhu2023minigpt4}
Deyao Zhu, Jun Chen, Xiaoqian Shen, Xiang Li, and Mohamed Elhoseiny. 2023.
\newblock \href {https://arxiv.org/abs/2304.10592} {Minigpt-4: Enhancing vision-language understanding with advanced large language models}.
\newblock \emph{Preprint}, arXiv:2304.10592.

\end{thebibliography}

% \appendix

% \section{Example Appendix}
% \label{sec:appendix}

% This is an appendix.

\end{document}